\documentclass[a4paper,11pt]{article}

\usepackage{fourier}
\usepackage{color}
\usepackage{graphicx}
\usepackage{url}
\usepackage[affil-it]{authblk}
\usepackage{amsmath}
\usepackage{wrapfig}

\usepackage[T1]{fontenc}
\usepackage{times}
\usepackage[normalem]{ulem}
\usepackage[table,xcdraw]{xcolor}
\usepackage[square]{natbib}
\usepackage{booktabs}

\usepackage{multirow}
\usepackage{hhline}
\usepackage{caption}
\usepackage{subcaption}
\usepackage{float}

\begin{document}

\title{Extracting Pasture Phenotype and Biomass Percentages using Weakly Supervised Multi-target Deep Learning on a Small Dataset }

\author[1]{Badri Narayanan}
\author[1]{Mohamed Saadeldin}
\author[2]{Paul Albert}
\author[2]{Kevin McGuinness}
\author[1]{Brian Mac Namee}
\affil[1]{School of Computer Science, University College Dublin.}
\affil[2]{Insight Centre for Data Analytics, Dublin City University.}
\date{07-07-2020}
\maketitle
\thispagestyle{empty}

\begin{abstract}
The dairy industry uses clover and grass as fodder for cows. Accurate estimation of grass and clover biomass yield enables smart decisions in optimizing fertilization and seeding density, resulting in increased productivity and positive environmental impact. Grass and clover are usually planted together, since clover is a nitrogen-fixing plant that brings nutrients to the soil. Adjusting the right percentages of clover and grass in a field reduces the need for external fertilization. Existing approaches for estimating the grass-clover composition of a field are expensive and time consuming---random samples of the pasture are clipped and then the components are physically separated to weigh and calculate percentages of dry grass, clover and weeds in each sample. There is growing interest in developing novel deep learning based approaches to non-destructively extract pasture phenotype indicators and biomass yield predictions of different plant species from agricultural imagery collected from the field. Providing these indicators and predictions  from images alone remains a significant challenge.  Heavy occlusions in the dense mixture of grass, clover and weeds make it difficult to estimate each component accurately. Moreover, although supervised deep learning models perform well with large datasets, it is tedious to acquire large and diverse collections of field images with precise ground truth for different biomass yields. In this paper, we demonstrate that applying data augmentation and transfer learning is effective in predicting multi-target biomass percentages of different plant species, even with a small training dataset. The scheme proposed in this paper used a training set of only 261 images and provided predictions of biomass percentages of grass, clover, white clover, red clover, and weeds with mean absolute error (MAE) of 6.77\%, 6.92\%, 6.21\%, 6.89\%, and 4.80\% respectively. Evaluation and testing were performed on a publicly available dataset provided by the Biomass Prediction Challenge \citep{Skovsen_2019_CVPR_Workshops}. These results lay the foundation for our next set of experiments with semi-supervised learning to improve the benchmarks and will further the quest to identify phenotype characteristics from imagery in a non-destructive way.
\end{abstract}
\textbf{Keywords:} Computer vision, deep learning, transfer learning, smart agriculture, data augmentation, weak supervision

\section{Introduction}
The dairy industry uses clover and grass as fodder for cows. Grass and clover are grown together in fields to improve the consistency of high-quality biomass yield and to reduce the need for external fertilizers. Accurate estimation of the dry biomass percentages of grass and clover species in fields is very important for determining optimal seeding density, fertilizer application and elimination of weeds. Conventionally, this has been done manually by clipping random sample areas in the field and sending it to the lab to visually identify, separate and weigh different species. This manual approach is laborious and time consuming, and gives rise to the need for a more efficient and non-destructive approach for biomass estimation. Machine learning approaches that directly estimate dry biomass percentages from images have huge potential for addressing this need. The task, however, is challenging. Two types of clover, red and white, are often grown together. Red clover has a shorter life cycle than white clover, but the two species are visually similar and difficult to distinguish \citep{skovsen2018predicting}. Moreover, in addition to grass and clover, there is usually presence of undesirable weeds too. The choice of an appropriate machine learning solution should focus on accurately discriminating between the two clover species while accurately predicting the grass-clover-weeds mix ratio. 

Deep learning models based on Convolutional Neural Networks (CNNs) have achieved impressive performance across various computer vision applications. This paper proposes a deep learning model that predicts biomass percentages of grass, white clover, red clover, and weeds from canopy view images collected from farms. The Grass Clover Image dataset from the Biomass Prediction Challenge \citep{Skovsen_2019_CVPR_Workshops} is used to develop this approach. This dataset poses unique challenges in the form of a multi-target prediction problem (a distribution of biomass yield across species types must be prdicted) and partial labelling with missing values of dry white and red clover in many examples. The training data also consists of just 261 labeled images, which is very small for training a deep model. CNNs can generalize efficiently when trained on large datasets, whereas small datasets pose an optimization problem, thus needing numerous iterations that invariably result in overfitting. 

In this paper, we demonstrate an improvement in the state-of-art techniques for biomass prediction by providing an end-to-end approach from pixels to biomass directly from real images. This is in contrast to prevalent multi-step approaches that train on a large amount of synthetic images and a small number of labelled real images. We adapt a VGG-16 model pretrained on Imagenet to the multi-target regression problem of predicting biomass percentages. This adaptation, in combination with data imputation for missing values and weak supervision through differential sample weights for weak labels, does a good job of extracting features. We avoided overfitting through model regularisation and data augmentation to increase the number of training samples, that resulted in improved prediction of biomass for the small dataset. 

The remainder of the paper proceeds as follows. Section \ref{sec:relatedWork} provides a review of current methods for pasture phenotype estimation, specifically those that apply deep learning techniques. Previous work in multi-target learning for regression problems, transfer learning and weak supervision is also highlighted. In Section 3, we explain the biomass percentages estimation process with the proposed adaptation of transfer learning and weak supervision. The section also covers a sequence of data pre-processing steps required to address the limitations of the dataset. Section 4 describes an evaluation experiment and discusses its results. Finally, in Section 5 we present our conclusions and indicators to our future research.

\section{Related work}\label{sec:relatedWork}
In this section, we review applications of deep learning in smart agriculture, as well as existing work on multi-target regression, transfer learning, and weak supervision. 

\subsection{Deep learning for biomass estimation}
\cite{kamilaris_deep_2018} present an extensive survey of  applications of deep-learning-based computer vision in smart agriculture. These applications include weed identification, land cover classification, plant recognition, fruit counting, and crop type classification, as well as biomass estimation. 
\cite{larsen2018autonomous} and \cite{skovsen2018predicting} used aFully Convolutional Network (FCN) architecture \citep{long2015fully} for semantic segmentation to identify plant species in images at a pixel level. \citeauthor{larsen2018autonomous}~use images captured by unmanned aerial vehicles from two different farms in Denmark to provide pixel coverage of species in an image. Whereas \citeauthor{skovsen2018predicting}~provide a two step approach of using 2 FCNs to classify pixels as grass, clover, weeds and soil in a large set of synthetic images, and further use these models on real images with biomass ground truth to predict biomass from the pixel percentages of the individual components through a regression model.


\subsection{Multi-target regression}
In a survey of multi-output learning, \cite{xu2019survey} describe  multi-target regression as a type of multi-output learning that predicts multiple real-valued outputs from a set of input features for an example instance. By simultaneously predicting all the targets, the algorithm captures the inherent relationships between the targets themselves in addition to their individual relationships with the input features. According to \cite{borchani2015survey}, capturing these relationships in the predictions will appropriately reflect real-world problems. 

\subsection{Transfer learning and weak supervision}

Transfer learning between task domains reduces the effort to label training data for the target domain by transferring knowledge from a pre-trained model on a large dataset from a different domain. \cite{pan_survey_2010} explain the different situations where the source and target domains and tasks are same or different and accordingly describe the three categories of transfer learning: \textit{inductive transfer learning, transductive transfer learning, and unsupervised transfer learning}. Additionally, they describe "what to transfer" through 4 different approaches: \textit{instance transfer, feature representation transfer, parameter transfer, and relational knowledge transfer.}  The \textit{inductive transfer} setting where the source and target tasks are different, irrespective of the difference in the domains, resembles our problem. In the problem addressed by this paper, the visual recognition (classification) task of VGG-16 \citep{simonyan2014very} approach using ImageNet (source domain) is different to the biomass prediction (regression) task with farm images (target domain). Our approach focused on learning latent features in the farm images by transferring feature representation knowledge from VGG-16, followed by a non-linear regression to predict the biomass from the learnt features. 

Small datasets that have missing values can benefit from weakly supervised learning \citep{zhou2018brief}. Weak supervision can be of three types: \textit{incomplete, inexact, and inaccurate}. Incomplete supervision refers to generating weak labels for a large subset of data that have missing labels because only a small subset can be hand-annotated by human experts. Inexact supervision arises in a scenario where fine-grained labels are difficult to annotate and the dataset has coarse-grained labels only. Inaccurate supervision is the case where the ground truth is largely expected to be imperfect because of various reasons like human error, crowdsourcing to obtain labels, or difficulties in recognizing and categorizing. 

In the next section we describe an approach to estimating biomass percentages from images of grass (a multi-target regression problem) using a CNN. To train this  CNN we utilise transfer learning and weak supervison as well as data  augmentation techniques.

\section{Estimating biomass percentages}
This paper addresses the problem of estimating biomass yield of grass, clover (red and white) and weeds at different seasons of crop growth directly from real farm imagery. In the process, we tackle some unique challenges. Firstly, we are predicting multiple targets with an overall percentage distribution and a further percentage distribution of sub-targets. Secondly, we deal with a small dataset of real farm images that has missing values. We evaluated and tested our proposed approach the publicly available Grass Clover Image Dataset for the Biomass Prediction Challenge \citep{Skovsen_2019_CVPR_Workshops}. 

 \cite{Skovsen_2019_CVPR_Workshops} describe a baseline 2-step approach to this challenge. The first step trains 2 FCN semantic segmentation models initialized with pretrained weights from VGG16. The FCN models are trained using 1720 synthetic images. The first FCN classifies the pixels in the images into grass, clover, weeds, and soil;  while the second FCN identifies red and white clover in them. Using 261 real high-resolution farm images with biomass labels a linear regression model is trained to predict the grass-white clover-red clover-weed dry biomass percentages from the the percentages of pixels of each type identified in the semantic segmentation performed by the two FCNs. 
The Biomass Prediction Challenge stipulates two metrics for each predicted category, the root mean square error (RMSE) and mean absolute error (MAE) to evaluate the model's performance.

\subsection{Dataset}
The Biomass Prediction Dataset has 261 training images captured from 3 different dairy farms, with their corresponding biomass compositions, expressed in actual and percentage terms. 174 unlabelled test images are provided along with with their harvest season for the contest evaluation. Each image is taken over a square frame of vegetation measuring $0.5m \times 0.5m$, at ground sampling distances (GSD) of 4-8 px mm\textsuperscript{-1}. The image sizes range approximately within $1800 \times 1800$ px to $3000 \times 3000$ px. 

The metadata associated with each image in the dataset is categorised as basic, semi-advanced or advanced, and the biomass weights are measured over different harvest seasons. The semi-advanced category corresponds to data collected in the 1\textsuperscript{st} seasonal harvest, and the advanced category refers to data collected in the other seasonal harvests. The dry biomass percentages of grass, clover and weeds comprise 100\% of the dry biomass for all the 261 training examples. The semi-advanced and advanced categories, totalling 157 examples, additionally reflect the break up for the subspecies of white and red clovers. In this study we did not use seasonal harvest as a distinguishing feature, and so we have grouped semi-advanced and advanced labels together under a single category (advanced), as reflected in Table \ref{tab:labels}.

\begin{table}[hbt!]
\centering
\begin{tabular}{@{}lrrr@{}}
\toprule
\begin{tabular}[c]{@{}l@{}}Seasonal \\ Harvest No\end{tabular} & Basic & Advanced & Grand Total \\ \midrule
1     & 23  & 37  & 60  \\
2     & 33  & 56  & 89  \\
3     & 25  & 35  & 60  \\
4     & 23  & 29  & 52  \\
\addlinespace
Total & 104 & 157 & 261 \\ \bottomrule
\end{tabular}
\caption{Data distribution and labeling.
Semi-advanced and Advanced label types have been grouped together under Advanced}
\label{tab:labels}
\end{table}


\subsection{Data imputation for missing label values}
Supervised learning requires datasets with complete labels for all data examples in the training set. The dataset, as explained above, is missing label values for the subspecies of white and red clover for all the 104 examples in the basic category, although the total clover biomass percentage is provided. This is a unique problem in a supervised learning setting where sub parts of the ground truth is unavailable and needs to be approximated in order to allow all data examples to be used for training.  

\cite{gelman2006data} describe different imputation techniques for estimating missing values using other available values. Three alternative methods, specifically multiple regression, mean and median value imputations were evaluated in this work for white and red clover label imputation. Models were trained using labels generated from each of the three alternative imputation methods for comparison of test results. 

The deterministic regression imputation technique ignores the error term and predicts exact values for the missing cells from the observed data for the corresponding variables. Since there are multiple variables with missing values, regression cannot be directly applied. To work around this, we initially applied random imputation for the missing cells, followed by deterministic regression imputation multiple times. We included the categorical variables related to the harvest season as predictors for the regression. 

In the alternative option of imputation using mean values, dry white clover and dry red clover were initially expressed as fractions of dry clover from Advanced categories. The mean values of these 2 fractions were then applied on the dry clover biomass of the basic category examples to calculate the respective proportions. The resulting proportions were then expressed as percentages of total biomass to get the imputed label values for white and red clover percentages respectively. We also explored imputing median values instead of mean as the third alternative. 

\subsection{Data augmentation}
Training deep neural networks on small datasets cause optimization issues and overfitting. The network does not encounter sufficient example images to learn enough features for generalisation. In such cases, \cite{krizhevsky2012imagenet} suggest that image augmentation can enhance the network's generalisation performance by artificially increasing the number of examples in the dataset through label-preserving transformations. In an earlier work, describing some good practices that can be applied to visual document analysis, \cite{simard2003best} explain that when working with datasets containing a small number of images, if transformation invariance properties are inherent in the image parameters, the generalisation performance of a network will improve when we feed the network with additional transformed data. The larger number of images enable the network to learn these invariances better. 

\begin{figure*}[h!]
\begin{subfigure}{0.15\textwidth}
  \centering
  Original images from the dataset
\end{subfigure}
    \hfill
\begin{subfigure}{0.2\textwidth}
  \centering
  \includegraphics[width=\linewidth]{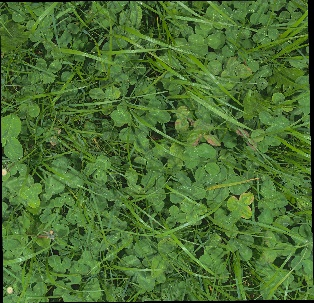}  
  \label{fig_original:orig-first}
\end{subfigure}
    \hfill
\begin{subfigure}{0.2\textwidth}
  \centering
  \includegraphics[width=\linewidth]{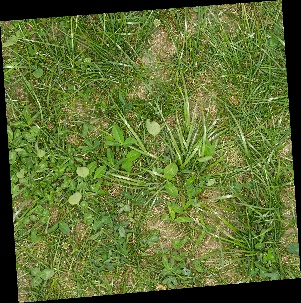}  
  \label{fig_original:orig-second}
\end{subfigure}
    \hfill
\begin{subfigure}{0.2\textwidth}
  \centering
  \includegraphics[width=\linewidth]{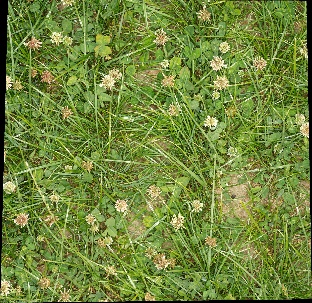}  
  \label{fig_original:orig-third}
\end{subfigure}
    \hfill
\begin{subfigure}{0.2\textwidth}
  \centering
  \includegraphics[width=\linewidth]{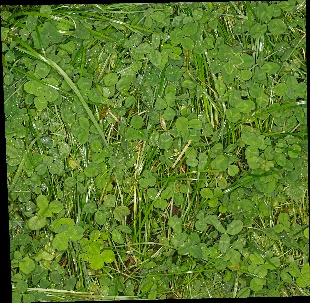}  
  \label{fig_original:orig-fourth}
\end{subfigure}
\bigskip
\begin{subfigure}{0.15\textwidth}
  \centering
  Examples of corresponding randomly augmented images
\end{subfigure}
    \hfill
\begin{subfigure}{0.2\textwidth}
  \centering
  \includegraphics[width=\linewidth]{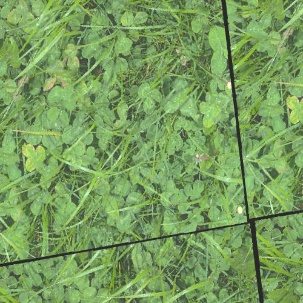}  
  \label{fig_trans:trans-1}
\end{subfigure}
    \hfill
\begin{subfigure}{0.2\textwidth}
  \centering
  \includegraphics[width=\linewidth]{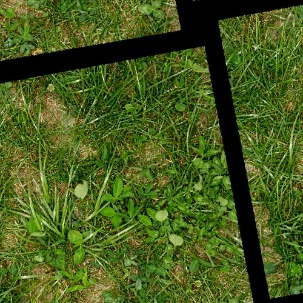}  
  \label{fig_trans:trans-2}
\end{subfigure}
    \hfill
\begin{subfigure}{0.2\textwidth}
  \centering
  \includegraphics[width=\linewidth]{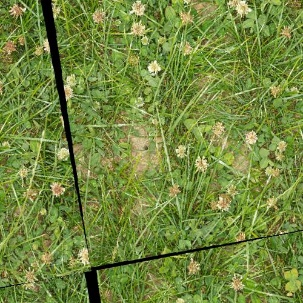}  
  \label{fig_trans:trans-3}
\end{subfigure}
    \hfill
\begin{subfigure}{0.2\textwidth}
  \centering
  \includegraphics[width=\linewidth]{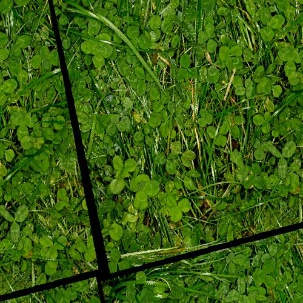}  
  \label{fig_trans:trans-4}
\end{subfigure}
\captionsetup{justification=centering}
\caption{Image augmentation samples. The augmented images are resized to $500\times500$ px. }
\label{fig:fig_grass}
\end{figure*}

For each image in our dataset, we fed the network with 10 modified training examples that are run-time random transformations of the original example image. Some sample transformations are shown in Figure \ref{fig:fig_grass}. All images were reduced to a standard size of $500\times500$ px, as against a higher resolution of $1200\times1200$ px used in the baseline benchmark in the competition. We employed a combination of operations for random transformations that includes rotation range of 15 degrees, zoom range of 15\%, 20\% width / height shift, shear range of 15\%, horizontal flips, and a channel shift range of 50. To prevent loss in the image from rotation and shifts, we used the wrap function for the fill\_mode parameter. 

A cursory inspection of random images revealed that the grass-clover distribution is not homogeneous within an image. Since the ground truth data points cannot be proportionally divided, we avoided splitting the images into multiple smaller images. A total of 2,090 run-time augmented images per epoch constituted the final training dataset, increasing the training size by 10x.

\subsection{Weak supervision}
Approximations in the training labels through data imputation to account for scarcity, introduce noise in the ground truth and implies weak labels for certain examples. Such examples are given lesser importance in training by lowering their weights in the calculation of the loss function. This \textit{Incomplete} form of weak supervision \citep{zhou2018brief} was adopted when training the model with imputed weak labels since they have a higher degree of uncertainty and are less reliable in training. We iterated with different options for the differential sample weights and determined that 1:1.5 for basic vs advanced gave the best results.

\subsection{Model architecture}
Transfer learning was employed for feature extraction by initializing the CNN with pre-trained weights from the the well known VGG-16 architecture trained on the ImageNet dataset. We retained the convolutional layers and dropped the last two dense layers that were originally designed for a classification task. The weights of the convolutional layers were made non-trainable. This was to facilitate detection of feature representations by the original architecture in the target dataset. The convolutional layers were followed by 2 dense layers of 4,096 and 256 neurons each, with ReLU activations and `random\_uniform' kernel initialization. Each dense layer was accompanied by a batch normalization. The architecture was completed with an output layer of 4 neurons and a softmax activation to predict the percentages for grass, white clover, red clover and weeds. 

The 261 images in the dataset were divided into a training set comprising of  209 images and a validation set with 52 images, which were kept standard across the different models for performance comparison. The 209 images in the training set were subjected to 10 random data augmentations with size $500\times500$ px at run-time, generating 2,090 different training images per epoch with a mini-batch size of 8. The network was thus never exposed to the original images. The network was trained for 100 epochs using Adam optimizer with a learning rate 10\textsuperscript{-3} and weight decay 10\textsuperscript{-3} / 200. The best weights based on the least validation loss across epochs were stored and used for predictions on the 174 unlabelled images in the test set for competition submissions. Root Mean Squared Error (RMSE) was used as the loss function.

\section{Results}
The model was trained three times with different sets of imputed labels obtained through mean, median and regression imputations. The results of the validation metrics RMSE and MAE based on each model's best weights for the least validation loss across epochs are compared for the three imputation techniques in Table~\ref{tab:val-results-table}. While it is clear that the median imputation technique yields the best overall result with the lowest MAE of 5.55, we find that the mean imputation technique performs the best in differentiating the white and red clovers with MAE scores of 5.99 and 5.63 respectively, and the overall MAE of 5.64 is comparable with the median technique. Ability to better discriminate the two subspecies of clover is a key objective.

\begin{table}[ht]
\small
\centering
\resizebox{\textwidth}{!}{\begin{tabular}{|l|c|c|c|c|c|c|c|c|c|c|c|c|}
\hline
\rowcolor[HTML]{C0C0C0} 
 &
  \multicolumn{2}{c|}{\cellcolor[HTML]{C0C0C0}Grass} &
  \multicolumn{2}{c|}{\cellcolor[HTML]{C0C0C0}Clover} &
  \multicolumn{2}{c|}{\cellcolor[HTML]{C0C0C0}White Clover} &
  \multicolumn{2}{c|}{\cellcolor[HTML]{C0C0C0}Red Clover} &
  \multicolumn{2}{c|}{\cellcolor[HTML]{C0C0C0}Weeds} &
  \multicolumn{2}{c|}{\cellcolor[HTML]{C0C0C0}Overall} \\ \hline
\rowcolor[HTML]{FFFFC7} 
Data Imputation technique & RMSE & MAE  & RMSE & MAE  & RMSE & MAE  & RMSE & MAE  & RMSE & MAE  & \cellcolor[HTML]{FFFFC7}RMSE & \cellcolor[HTML]{FFFFC7}MAE \\ \hline
Regression &
  {\color[HTML]{000000} 8.14} &
  {\color[HTML]{000000} 6.84} &
  {\color[HTML]{000000} 8.55} &
  {\color[HTML]{000000} 6.92} &
  {\color[HTML]{000000} 8.24} &
  {\color[HTML]{000000} 6.44} &
  {\color[HTML]{000000} 9.33} &
  {\color[HTML]{000000} 6.80} &
  {\color[HTML]{000000} 5.51} &
  {\color[HTML]{000000} 3.95} &
  8.06 &
  6.19 \\ \hline
Mean                      & 8.00 & 6.21 & 7.69 & 6.16 & 7.44 & 5.99 & 7.33 & 5.63 & 5.68 & 4.20 & 7.27                         & 5.64                        \\ \hline
Median                    & 7.45 & 5.95 & 7.26 & 6.04 & 8.05 & 6.2  & 8.4  & 6.12 & 4.57 & 3.42 & 7.27                         & 5.55                        \\ \hline
\end{tabular}}
\caption{Comparison of validation metrics RMSE and MAE for each biomass component as well as the overall performance across different imputation techniques.
}
\label{tab:val-results-table}
\end{table}

The model weights that provided the lowest loss error on the validation set during training were saved and used to generate output predictions from the model on the test-set. Test results obtained from the model for each of the three different runs were submitted to the challenge website for evaluation against ground truth. The challenge organisers evaluate the submitted results independently. RMSE results for (grass, clover, white clover, red clover, and weeds) obtained on the test set using the proposed model with different label imputation schemes are summarized in Table \ref{tab:results_RMSE}. Similarly, MAE results are provided in Table \ref{tab:results_MAE}.

\begin{table}[htb]
\small
\centering
\caption{RMSE results on the test dataset for different imputation schemes compared to baseline }
\begin{tabular}{|l|c|c|c|c|c|c|c|c|c|c|}
\hline

\rowcolor[HTML]{BFBFBF} 
  Data Imputation &
  Grass &
  
  Clover &
  
  White Clover &
 
  Red Clover &
  
  Weeds 
   \\ \hline
\rowcolor[HTML]{FFFFC7} 
Baseline &
  9.05 &
  
  9.91 &
  
  9.51 &
  
  \textbf{6.68} &
  
  \textbf{6.49} 
   \\ \hline
Regression &
  8.98 &
  
  10.03 &
  
  8.78 &
  
  10.46 &
  
  6.86 
  \\ \hline
Mean &
  \textbf{8.64} &
  
  \textbf{8.73} &
  
  8.16 &
  
  10.11 &
 
  6.95 
   \\ \hline
Median &
  8.67 &
  
  9.93 &
  
  \textbf{8.09} &
  
  9.87 &
  
  7.73 
   \\ \hline
\end{tabular}
\captionsetup{justification=centering}
\label{tab:results_RMSE}
\end{table}

\begin{table}[ht]
\small
\centering
\caption{MAE results on test data for different imputation schemes compared to baseline }
\begin{tabular}{|l|c|c|c|c|c|c|c|c|c|c|}
\hline

\rowcolor[HTML]{BFBFBF} 
  Data Imputation &
  Grass &
  
  Clover &
  
  White Clover &
 
  Red Clover &
  
  Weeds 
   \\ \hline
\rowcolor[HTML]{FFFFC7} 
Baseline &
  
  6.85 &
  
  7.82 &
  
  7.61 &
  
  \textbf{4.84} &
  
  \textbf{4.61} \\ \hline
Regression &
  
  6.96 &
  
  7.94 &
  
  6.84 &
  
  7.70 &
  
  4.80 \\ \hline
Mean &
  
  \textbf{6.77} &
  
  \textbf{6.92} &
  
 \textbf{6.21} &
  
  7.74 &
  
  5.02 \\ \hline
Median &
  
  6.89 &
 
  8.06 &
  
  6.23 &
  
  6.89 &
 
  5.95 \\ \hline
\end{tabular}
\captionsetup{justification=centering}
\label{tab:results_MAE}
\end{table}

In general the mean imputation scheme for labels provide better results compared to other schemes. Comparing the proposed model results to the baseline the model obtains lower error for the percentages of grass, clover, and white clover. The proposed model error results for weeds percentages are slightly higher than the baseline. Regarding red clover higher error compared to baseline it is very tricky to distinguish between White and red clover since they look quite similar. Further tuning or even model structure modifications might be required to improve the differentiation between white and red clover.

\section{Conclusion}
This paper describes a deep learning approach to estimating biomass percentages of different plant types grown together in the field. Traditional approaches for estimating biomass percentages are destructive,  expensive and time consuming, since they require random samples of the field to be clipped and then manually separated and weighed. In this paper, a prediction approach based on transfer learning utilizing a VGG-16 model pre-trained on Imagenet, while adapting the final dense layers to perform multi-target regression is presented. The proposed scheme benefits from data augmentation, label imputation, and weak supervision to achieve best prediction outputs. The proposed model and training scheme was tested and evaluated on the the Grass Clover Dataset provided by the Biomass Prediction Challenge \cite{Skovsen_2019_CVPR_Workshops}.  Though the training set consists of only 261 images, the proposed model achieved predictions of biomass percentages for grass, clover, white clover, red clover, and weeds with Mean Absolute Error (MAE) of 6.77\%, 6.92\%, 6.21\%, 6.89\%, and 4.80\% respectively. The proposed model predictions outperform the baseline provided in Biomass Prediction Challenge \cite{Skovsen_2019_CVPR_Workshops} as it provide lower error for grass, clover, and white clover biomass percentages. The obtained prediction error for red clover and weeds is higher than baseline with a small margin for weeds. The initial choice of VGG16 for transfer learning is arbitrary and provides an early proof of concept. In the future, we plan to experiment with other networks such as ResNet, wResNet, or deeper VGG architectures. We will also investigate more sophisticated weak supervision and semi-supervised learning schemes that take advantage of unlabelled data.

\newpage

\section*{Acknowledgments}
This publication has emanated from research conducted with the financial support of Science Foundation Ireland (SFI) and the Department of Agriculture, Food and Marine on behalf of the Government of Ireland under Grant Number [16/RC/3835] - VistaMilk.

\bibliographystyle{apalike}

\bibliography{imvip}

\end{document}